# A Temporal Neuro-Fuzzy Monitoring System to Manufacturing Systems

Rafik Mahdaoui [1,2], Leila Hayet Mouss[1], Mohamed Djamel Mouss [1], Ouahiba Chouhal [1,2]

1 Laboratoire d'Automatique et Productique (LAP) Université de Batna,
Rue Chahid Boukhlouf 05000 Batna, Algérie
1,2  Centre universitaire Khenchela  Algérie,
Route de Batna BP:1252, El Houria, 40004 Khenchela Algérie

**Abstract**

Fault diagnosis and failure prognosis are essential techniques in improving the safety of many manufacturing systems. Therefore, on-line fault detection and isolation is one of the most important tasks in safety-critical and intelligent control systems.

Computational intelligence techniques are being investigated as extension of the traditional fault diagnosis methods. This paper discusses the Temporal Neuro-Fuzzy Systems (TNFS) fault diagnosis within an application study of a manufacturing system. The key issues of finding a suitable structure for detecting and isolating ten realistic actuator faults are described. Within this framework, data-processing interactive software of simulation baptized NEFDIAG (NEuro Fuzzy DIAGnosis) version 1.0 is developed.

This software devoted primarily to creation, training and test of a classification Neuro-Fuzzy system of industrial process failures. NEFDIAG can be represented like a special type of fuzzy perceptron, with three layers used to classify patterns and failures. The system selected is the workshop of SCIMAT clinker, cement factory in Algeria.

## 1. Introduction

Several methods have been proposed in order to solve the fault detection and fault diagnosis problems. The most commonly employed solution approaches for fault diagnosis system include (a) model-based, (b) knowledge-based, and (c) pattern recognition-based approaches. Generally, analytical model-based  methods can be designed in order to minimize the effect of unknown disturbance and perform the consistent sensitivity analysis. Knowledge-based methods are used when there is a lot of experience but not enough details to develop accurate quantitative models. Pattern recognition methods are applicable to a wide variety of systems and exhibit real-time characteristics. [8]. Therefore the human expert in his mission of diagnosing the cause of a failure of a whole system, uses quantitative or qualitative information.  On another side, in spite of the results largely surprising obtained by the ANN in monitoring and precisely in diagnosis they remain even enough far from equalizing the sensory capacities and of reasoning human being.  Fuzzy logic makes another very effective axis in industrial diagnosis.

Also, can we replace the human expert for automating the task of diagnosis by using the Neuro-fuzzy approach? In addition, how did the human expert gather all relevant information and permit him to make their decision?  Our objective consists of the following: making an association (adaptation) between the techniques of fuzzy logic and the temporal neural networks techniques (Neuro-fuzzy system), choosing the types of neural networks, determining the fuzzy rules, and finally determining the structure of the temporal Neuro-Fuzzy system to maximize the automation of the diagnosis task.

In order to achieve this goal we organize this article into three parts. The first part presents principal architectures of diagnosis an prognosis methods and principles for Temporal Neuro-Fuzzy systems operation and their applications (sections 2 and 3).The second part is dedicated to the workshop of clinker of cement factory (Section 4). Lastly, in the third part we propose a Neuro-Fuzzy system for system of production diagnosis. Machine Fault Prognosis

The literatures of prognosis are much smaller in comparison with those of fault diagnosis. The most obvious and normally used prognosis is to use the given current and past machine condition to predict how much time is left before a failure occurs. The time left before observing a failure is usually called remaining useful life (RUL). In order to predict the RUL, data of the fault propagation process and/or the data of the failure mechanism must be available. The fault propagation process is usually tracked by a trending or forecasting model for certain condition variables. There are two ways in describing the failure mechanism. The first one assumes that failure only depends on the condition variables, which reflect the actual fault level, and the predetermined boundary (figure 1) .

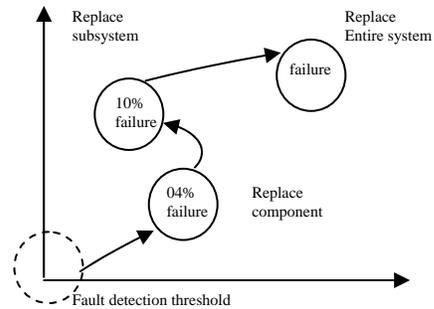

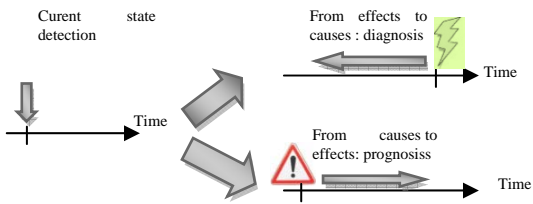

Figure 1. detection ,diagnosis and prognosis- the phenomenological aspect

Figure 2. Prognosis technical approaches

The definition of failure is simply defined that the failure occurs when the fault reaches a predetermined level. The second one builds a model for the failure mechanism using available historical data. In this case, different definitions of failure can be defined as follows: (a) an event that the machine is operating at an unsatisfactory level; or (b) it can be a functional failure when the machine cannot perform its intended function at all; or  (c) it can be just a breakdown when the machine stops operating, etc.

The approaches to prognosis fall into three main categories: statistical approaches, model-based approaches, and data-driven based approaches.

Data-driven techniques are also known as data mining techniques or machine learning techniques. They utilize and require large amount of historical failure data to build a prognostic model that learns the system behavior. Among these techniques, artificial intelligence was regularly used because of its flexibility in generating appropriate model.

Several of the existing approaches used ANNs to model the systems and estimate the RUL. Zhang and Ganesan [14] used self-organizing neural networks for multivariable trending of the fault development to estimate the residual life of bearing system. Wang and Vachtsevanos [13] proposed an architecture for prognosis applied to industrial  chillers. Their prognostic model included dynamic wavelet neural networks, reinforcement learning, and genetic algorithms. This model was used to predict the failure growth of bearings based on the vibration signals. SOM and back propagation neural networks (BPNN) methods using vibration signals to predict the RUL of ball bearing were applied by Huang et al. in [12].

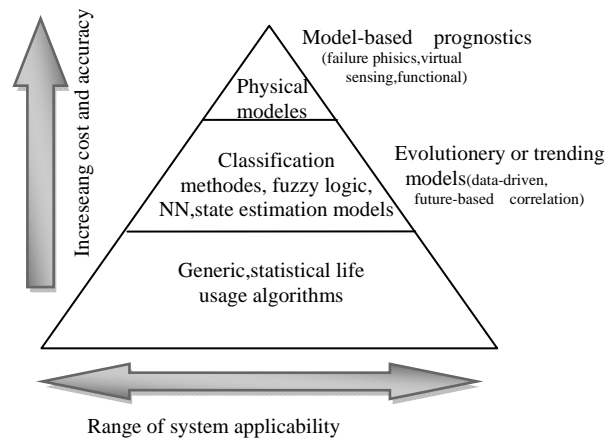

Figure 3. Prognosis technical approaches

Wang et al. [14] utilized and compared the results of two predictors, namely recurrent neural networks and ANFIS, to forecast the damage propagation trend of rotating machinery. In [15], Yam et al. applied a recurrent neural network for predicting the machine condition trend. Dong et al. [16] employed a grey model and a BPNN to predict the machine condition. Altogether, the data-driven techniques are the promising and effective techniques for machine condition prognosis.

## 2. Temporal Neuro-Fuzzy Systems

Fuzzy neural network (FNN) approach has become a powerful tool for solving real-world problems in the area of forecasting, identification, control, image recognition and others that are associated with high level of uncertainty [2,7,10,11,14,23,24,23]

The Neuro-fuzzy model combines, in a single framework, both numerical and symbolic knowledge about the process. Automatic linguistic rule extraction is a useful aspect of NF especially when little or no prior knowledge about the process is available [3]. For example, a NF model of a non-linear dynamical system can be identified from the empirical data.

This model can give us some insight about the on linearity and dynamical properties of the system.

The most common NF systems are based on two types of fuzzy models TSK [5] [7] combined with NN learning algorithms. TSK models use local linear models in the consequents, which are easier to interpret and can be used for control and fault diagnosis [23]. Mamdani models use fuzzy sets as consequents and therefore give a more qualitative description. Many Neuro-fuzzy structures have been successfully applied to a wide range of applications from industrial processes to financial systems, because of the ease of rule base design, linguistic modeling, and application to complex and uncertain systems, inherent non-linear nature, learning abilities, parallel processing and fault-tolerance abilities. However, successful implementation depends heavily on prior knowledge of the system and the empirical data [25].

Neuro-fuzzy networks by intrinsic nature can handle limited number of inputs. When the system to be identified is complex and has large number of inputs, the fuzzy rule base becomes large.

NF models usually identified from empirical data are not very transparent. Transparency accounts a more meaningful description of the process i.e less rules with appropriate membership functions. In ANFIS [2] a fixed structure with grid partition is used. Antecedent and consequent parameters are identified by a combination of least squares estimate and gradient based method, called hybrid learning rule. This method is fast and easy to implement for low dimension input spaces. It is more prone to lose the transparency and the local model accuracy because of the use of error back propagation that is a global and not locally nonlinear optimization procedure. One possible method to overcome this problem can be to find the antecedents & rules separately e.g. clustering and constrain the antecedents, and then apply optimization.

Hierarchical NF networks can be used to overcome the dimensionality problem by decomposing the system into a series of MISO and/or SISO systems called hierarchical systems [14]. The local rules use subsets of input spaces and are activated by higher level rules[12].

The criteria on which to build a NF model are based on the requirements for faults diagnosis and the system characteristics. The function of the NF model in the FDI scheme is also important i.e. Preprocessing data, Identification (Residual generation) or classification (Decision Making/Fault Isolation).

For example a NF model with high approximation capability and disturbance rejection is needed for identification so that the residuals are more accurate.

Whereas in the classification stage, a NF network with more transparency is required.

The following characteristics of NF models are important:

Approximation/Generalisation capabilities
transparency: Reasoning/use of prior knowledge /rules
Training Speed/ Processing speed
Complexity
Transformability: To be able to convert in other forms of NF models in order to provide different levels of transparency and approximation power.
Adaptive learning

Two most important characteristics are the generalising and reasoning capabilities. Depending on the application requirement, usually a compromise is made between the above two.

In order to implement this type of Neuro-Fuzzy Systems For Fault Diagnosis and Prognosis and exploited to diagnose of dedicated production system we have to propose data-processing software NEFDIAG (Neuro-Fuzzy Diagnosis).

The Takagi-Sugeno type fuzzy rules are discussed in detail in Subsection A. In Subsection B, the network structure of FENN is presented.

2.1 Temporal Fuzzy rules

Recently, more and more attention has paid to the Takagi-Sugeno type rules [9] in studies of fuzzy neural networks. This significant inference rule provides an analytic way of analyzing the stability of fuzzy control systems. If we combine the Takagi-Sugeno controllers together with the controlled system and use state-space equations to describe the whole system [10], we can get another type of rules to describe nonlinear systems as below:

Rule r: IF $X_1$ is $T^r_{x_1}$ AND ... ... AND $X_n$ is $T^r_{x_N}$ AND

$U_1$ is $T^r_{u_1}$ AND ... ... AND $U_M$ is $T^r_{U_M}$

**THEN** $\quad X = \dot{A}^r X + B^r U$

Where $X = [x_1 \ x_2 \ ..... \ x_n]^T$ is the inner is the inner state vector of the nonlinear system,

$U = [u_1 \ u_2 \ ..... \ u_n]^T$ is the input vector to the system, and N, M are the dimensions;

$T^r_{x_1}, T^r_{u_1}$ are linguistic terms (fuzzy sets) defining the conditions for $x_i$ and $u_j$ respectively, according to Rule r;

$A^r = (a^r_{ij})_{N*N}$ is a matrix of $N \times N$ and

$B^r = (b^r_{ij})_{N*M}$ Of $N \times M$

When considered in discrete time, such as modeling using a digital computer, we often use the discrete state-space equations instead of the continuous version. Concretely, the fuzzy rules become:

Rule r:
**IF** $X_1(t)$ is $T^r_{x_1}$ AND ... ... AND $X_n(t)$ is $T^r_{x_N}$ AND

$U_1(t)$ is $T^r_{u_1}$ AND ... ... AND $U_M(t)$ is $T^r_{U_M}$

**THEN** $\quad X(t+\dot{1}) = A^r X(t) + B^r U(t)$

Where $X = [x_1(t) \ x_2(t) \ ..... \ x_n(t)]^T$
is the discrete sample of state vector at discrete time t. In following discussion we shall use the latter form of rules.

In both forms, the output of the system is always defined as:
Y = CX ( or Y(t)= CX(t))        (1).
Where C= $(c_{ij})_{P \times X}$ is a matrix of PxN, and P is the dimension of output vector Y.

The fuzzy inference procedure is specified as below. First, we use multiplication as operation AND to get the firing strength of Rule r:

$$f_r = \prod_{i=1}^{N} \mu_{T^r_{xi}}[x_i(t)] \cdot \prod_{i=1}^{M} \mu_{T^r_{xi}}[u_i(t)] \quad (2)$$

Where $\mu_{T^r_{xi}}$ and $\mu_{T^r_{ui}}$ are the membership functions of $T^r_{xi}$ and $T^r_{u_i}$ respectively? After normalization of the firing strengths, we get (assuming R is the total number of rules)

$$S = \sum_{r=1,n}^{R} f_r \quad , h_r = f_r / S \quad (3)$$

Where S is the summation of firing strengths of all the rules, and $h_r$ is the normalized firing strength of Rule r. When the defuzzification is employed, we have

$X^r(t+1) = A^r X(t) + B^r U(t)$,

$$X(t+1) = \sum_{r=1}^{R} h_r X^r(t+1) \quad (4)$$

$$= \sum_{r=1}^{R} h_r [A^r X(t) + B^r U(t)]$$

$$= (\sum_{r=1}^{R} h_r A^r) X(t) + (\sum_{r=1}^{R} h_r B^r) U(t)$$

$$= AX(t) + BU(t)$$

Where $A = (\sum_{r=1}^{R} h_r A^r)$, $B = (\sum_{r=1}^{R} h_r B^r)$

Using equation (4), the system state transient equation, we can calculate the next state of system by current state and input.

### 2.2 The structure of temporal Neuro-Fuzzy System

The main idea of this model is to combine simple feed forward fussy systems to arbitrary hierarchical models.
The structure of recurrent Neuro-fuzzy systems is presented in figure 3:

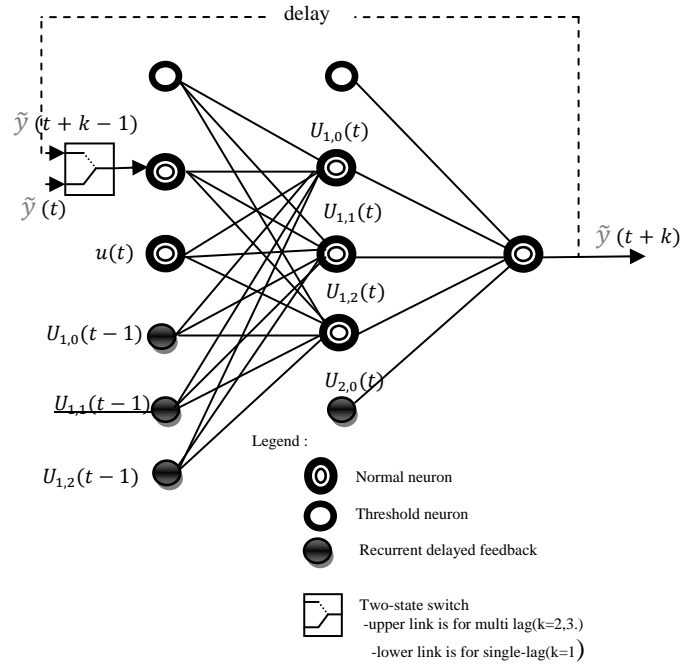

Fig 4. The structure of a simple TNFS

In this network, input nodes which accept the environment inputs and context nodes which copy the value of the state-space vector from layer 3 are all at layer 1 (the Input Layer). They represent the linguistic variables known as uj and xi in the fuzzy rules. Nodes at layer 2 act as the membership functions, translating the linguistic variables from layer 1 into their membership degrees. Since there may exist several terms for one linguistic variable, one node in layer 1 may have links to several nodes in layer 2, which is accordingly named as the term nodes. The number of nodes in the Rule Layer (layer 3) and the one of the fuzzy rules are the same - each node represents one fuzzy rule and calculates the firing strength of the rule using membership degrees from layer 2. The connections between layer 2 and layer 3 correspond with the antecedent of each fuzzy rule. Layer 4, as the Normalization Layer, simply does the normalization of the firing strengths. Then with the normalized firing strengths hr, rules are combined at layer 5, the Parameter Layer, where A and B become available. In the Linear System Layer, the 6th layer, current state vector X(t) and input vector U(t) are used to get the next state X(t +1), which is also fed back to the context nodes for fuzzy inference at time (t +1). The last layer is the Output Layer, multiplying X(t +1) with C to get Y(t +1) and outputting it.

Next we shall describe the feed forward procedure of TNFS by giving the detailed node functions of each layer, taking one node per layer as example. We shall use notations like $u_i^{[k]}$ to denote the i$^{th}$ input to the node in layer k, and o[k] the output of the node in layer k. Another issue to mention here is the initial values of the context nodes. Since TNFS is a recurrent network, the initial values are essential to the temporal output of the network. Usually they are preset to 0, as zero-state, but non-zero initial state is also needed for some particular case.

*Layer 1*. There is only one input to each node at layer 2. The Gaussian function is adopted here as the membership function:

$$o^{[1]} = e^{\frac{(u^{[1]} - c^r)^2}{2(s^r)^2}} \qquad (5)$$

where $c^r$ and $s^r$ give the center (mean) and width(variation) of the corresponding $u^{[1]}$ linguistic term of input u[2] in Rule r.

*Layer 2*. this layer has several nodes, one for figuring matrix A and the other for B. Though we can use many nodes to represent the components of A and B separately, it is more convenient to use matrices. So with a little specialty, its weights of links from layer 4 are matrices Ar (to node for A) and Br (to node for B). It is also fully connected with the previous layer. The functions of nodes for A and B are respectively.

$$for\ A\ \ o^{[2]} = \sum_{r=1}^{R} u_r^{[2]} A^r, for\ B\ \ o^{[2]} = \sum_{r=1}^{R} u_r^{[2]} B^r \qquad (6)$$

*Layer 3*. the Linear System Layer has only one node, which has all the outputs of layer 1 and layer 2 connected to it as inputs. Using matrix form of inputs and output, we have [see (3)]

$$o^{[3]} = AX + BU = o_{for\ A}^{[2]} o_{context}^{[1]} + o_{for\ B}^{[2]} o_{context}^{[1]}$$

So the output of layer 3 is $X(t + 1)$ in (4).

This proposed network structure implements the dynamic system combined by our discrete fuzzy rules and the structure of recurrent networks. With preset human knowledge, the network can do some tasks well. But it will do much better after learning rules from teaching examples. In the next section, a learning algorithm will be put forth to adjust the variable parameters in FENN, such as $c^r$, $s^r$, $A^r$, $B^r$, and $C$.

## 3. Proposed Architecture for Fault diagnosis and Prognosis

Faults are usually the main cause of loss of productivity in the process industry. This section uses a straightforward architecture to detect, isolate and identify faults.

One of the most important types of systems present in the process industry is workshop of SCIMAT clinker . A fault in a workshop of SCIMAT clinker may lead to a halt in production for long periods of time. Apart from these economic considerations faults may also have security implications. A fault in an actuator may endanger human lives, as in the case of a fault in an elevator's emergency brakes or in the stems position control system of a nuclear power plant. The design and performance testing of fault diagnosis systems for industrial process often requires a simulation model since the actual system is not available to generate normal and faulty operational data needed for design and testing, due to the economic and security reasons that they would imply.

Figure 5 shows a view and the schematics of a typical industrial industrial process of manufacture of cement. This installation belongs to cement factory of Ain-Touta (SCIMAT) ALGERIA. This cement factory have a capacity of 2.500.000 t/an " Two furnaces " is made up of several units which determine the various phases of the manufacturing process of cement. The workshop of cooking gathers two furnaces whose flow clinker is of 1560 t/h. The cement crushing includes two crushers of 100t/h each one. Forwarding of cement is carried out starting from two stations, for the trucks and another for the coaches.

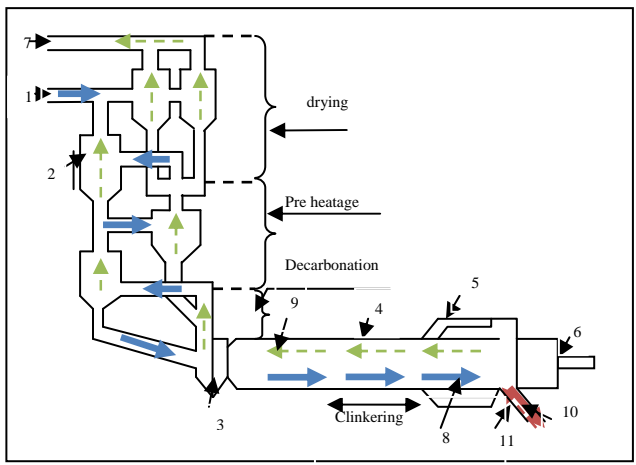

Fig 5. Workshop of SCIMAT clinker

### 3.1 Faults

The workshop of SCIMAT clinker may be affected by a number of faults. These faults are grouped into four major categories: heating tower faults, Kiln Cycling faults, cooler balloons faults and gas burner faults. Here only abrupt or incipient faults are considered.

This step has an objective of the identification of the dysfunctions which can influence the mission of the system. This analysis and recognition are largely facilitated using the structural and functional models of the installation. For the analysis of the dysfunctions we adopted the method for the analysis of the dysfunctions we adopted the method of Failure Modes and Effects Analysis and their Criticality (FMEAC).

While basing itself on the study carried out by [6], on the cooking workshop, we worked out an FMEAC by considering only the most critical modes of the failures (criticality >10), and for reasons of simplicity [46]. Therefore we have a Neuro-fuzzy system of 27 inputs and 11 outputs which were used to make a Prognosis of our system. The rules which are created with the system are knowledge a priori, a priori the base of rule. Each variable having an initial partition will be modified with the length of the phase of training (a number of sets fuzzy for each variable). The reasoning for the diagnosis and prognosis is described in the form of fuzzy rules inside our Neuro-fuzzy system.

Table 4.1 faults description

| Fault | Description | Inceptient/Abrupt |
|---|---|---|
| F1 | Chute de la jupe | I/A |
| F2 | bourrage | I/A |
| F3 | No break | I/A |
| F4 | Transporateur à auget | I/A |
| F5 | Presence anneaux | I |
| F6 | Mauvaise homogénéisation | I/A |
| F7 | Chute de croûtage | I/A |
| F8 | Atteinte des briques réfractaires | I |
| F9 | bourrage | I/A |
| F10 | Moteur ventilateur tirage | I/A |
| F11 | Courroies ventilateur tirage | I/A |

Our TNFS must have a number of inputs equal to the number of variables sensor signals providing the ability to extend the timing window used for this problem have 27 inputs nodes comprised of 11 sensors signals at 4 successive time points at steps of 10 minutes, resulting in a temporal window of 40 minutes for each sensor .

The TNFS provides 14 outputs representing the 14 possible classes (faults): 11 process faults, 3 sensor faults and normal state.

### 3.2 Training TNFS

To train the TNFS ,we used scenario for each of the 11 possible faults. The process was simulated for 120 minutes, with the faults starting to appear after 40 minutes of normal operation. So, we had 9 different positions of the temporal window (0-40 mins,10-50 mins, etc..), providing 342 input/output vector pairs for training.

NEFDIAG(Neuro-Fuzzy Diagonsis) is a data processing program for interactive simulation. The NEFDIAG development was carried out within LAP (University of Batna), was primarily dedicated to the creation, the training, and the test of a Neuro-Fuzzy system for the classification of the breakdowns of a dedicated industrial process. NEFDIAG models a fuzzy classifier Fr with a whole of classes C = {c1, c2...... cm}[45].

NEFDIAG makes it's training by a set of forms and each form will be affected (classified) using one of the preset classes. Next NEFDIAG generates the fuzzy rules by: evaluating of the data, optimizing the rules via training and using the fuzzy subset parameters, and partitioned the data into forms «characteristic» and classified with parameters of the data. NEFDIAG can be used to classify a new observation. The system can be represented in the form of fuzzy rules

**If** symptom1(t) is $A_1$  Symptom2(t-2) is $A_2$
Symptom3(t) is $A_3$  Symptom $_N$ (t-1) is $A_n$
**Then** the form ($x_1, x_2, x_3..., x_n$) belongs to class «fault i».

For example A1 A2 A3 An are linguistic terms represented by fuzzy sets. This characteristic will make it possible to complete the analyses on our data, and to use this knowledge to classify them. The training phase of the networks of artificial Neuro-Fuzzy systems makes it possible to determine or modify the parameters of the network in order to adopt a desired behavior. The stage of training is based on the decrease in the gradient of the average quadratic error made by network RNF[44].

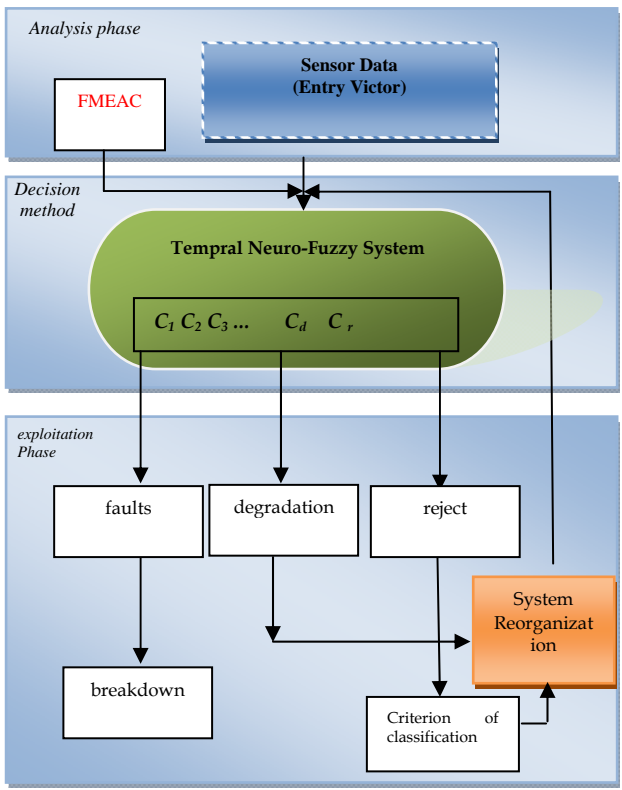

Fig. 6. The diagnosis by NEFDIAG.

The NEFDIAG system typically starts with a knowledge base comprised of a collection partial of the forms, and can refine it during the training. Alternatively NEFDIAG can start with an empty base of knowledge. The user must define the initial number of the functions of membership for partitioning the data input fields. And it is also necessary to specify the number K, which represents the maximum number of the neurons for the rules which will be created in the hidden layer. The principal steps of the training algorithm.

The data set used in this experiment contained 200 samples. Each data sample consisted of 27 features comprising the temperature and pressure measurements at various inlet and outlet points of the rotary kiln, as well as other important parameters as shown in Table 4.2. The heat transfer conditions were classified into two categories, i.e., the process of heat transfer was accomplished either efficiently or inefficiently.

From the database, there were 101 data samples (50.18%) that showed inefficient heat transfer condition, whereas 99 data samples (49.82%) showed efficient heat transfer condition in the rotary kiln. The data samples were equally divided into three subsets for training, prediction and test.

Table 4.2 input and output variables for the rule compiling

| Input var | Description |
|---|---|
| CO | CO in the first combustion chamber |
| Temp | Temp in the first combustion chamber |
| O2 | O2 in the second combustion chamber |
| RPM | Rotary kiln rotating RPM |
| Press | Pressure in the first combustion chamber |
| output var | Description |
| $\Delta Burner$ | Chang in burner heating power. i.e. burner(t)=burner(t-1)+ $\Delta Burner$(t) |
| $\Delta Air$ | Change in input air quantity ; i.e. Air(t)=air(t-1)+$\Delta Air$(t) |
| $\Delta IDFan$ | Change in induced fan inducing power i.e. IDFan(t)=IDFan(t-1)+$\Delta IDFan$(t) |

Usually, the structure of TNFS is determined by trial-and-error in advance for the reason that it is difficult to consider the balance between the number of rules and desired performance [20]. In this study, to determine the structure of TNFS, first we convert numeric data into information granules by fuzzy clustering. The number of clusters defines the number of fuzzy rules. By applying the fuzzy C-means clustering method [13,40] on the training data and checking the validity measure suggested in [13] it was identified that an adequate number of clusters is 4. Therefore 4 fuzzy rules were used for the basis for training and further refining. The clustering algorithm identified the following cluster centers for the presented data.

IF y(t-2) is A1 AND y(t-1) is B1 AND y(t) is C1 THEN y(t+1) is D1
IF y(t-2) is A2 AND y(t-1) is B2 AND y(t) is C2 THEN y(t+1) is D2
IF y(t-2) is A3 AND y(t-1) is B3 AND y(t) is C3 THEN y(t+1) is D3
IF y(t-2) is A4 AND y(t-1) is B4 AND y(t) is C4 THEN y(t+1) is D4

Initial fuzzy terms A1, A2, A3, A4 were created from the component y(t-2) of the cluster vectors 1, 2, 3, and 4, respectively. Similarly, terms B1, B2, B3, B4 – from y(t-1), C1, C2, C3, C4 – from y(t), and D1, D2, D3, D4 – from y(t+1). The terms A1, A2, ...,B1, B2, ..., C1, C2,...D1, D2, ... are described linguistically.

Figure 7 and 8 show the response of the normal model output and the real output from five to fifteen minutes prediction horizon and figure 9 to 10 show the response of the fault model output and the real output from three to seven minutes prediction horizon for test data.

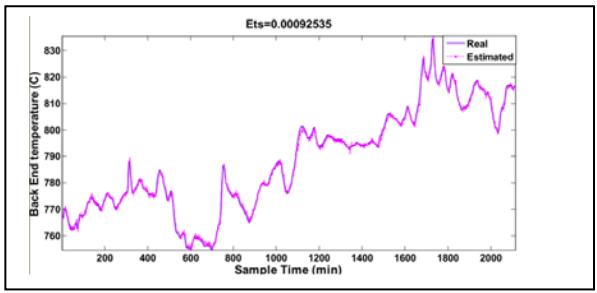

Fig. 7. Normal model with 5 min prediction horizon

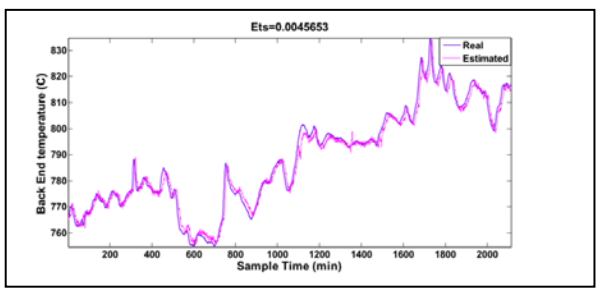

Fig. 8. Normal model with 10 min prediction horizon

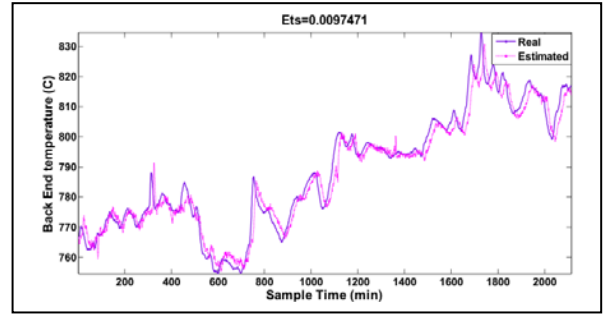

Fig. 9. Normal model with 15 min prediction horizon

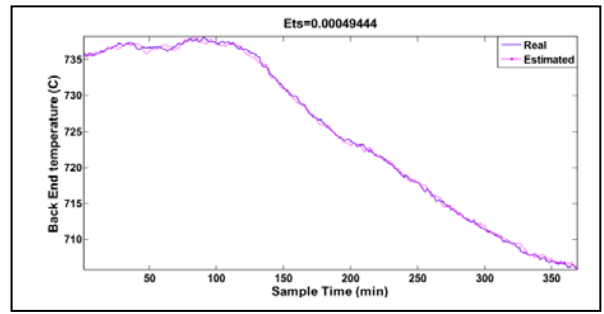

Fig.10 . Failure model with 10 or 15 min prediction

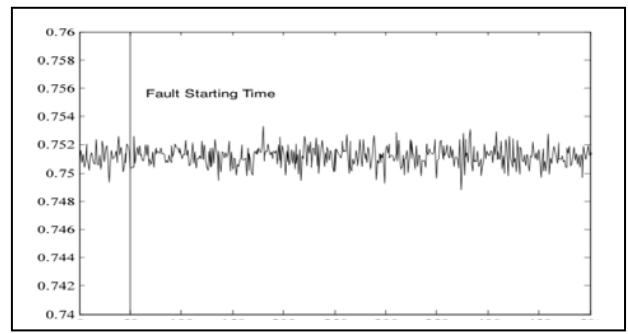

Fig. 11. Effect of incipient fault F10 on the Rotary kiln rotating RPM

## 4. Conclusion

The intelligent process operation aid system was developed to prevent the faults or errors in the process of manufacture of cement. This installation belongs to cement factory of Ain-Touta (SCIMAT) ALGERIA.

In order to do this work, the rule based and temporal Neuro-fuzzy system was implemented.

This TNFS was used for identification, prediction and detection of the fault process in the cement rotary kiln, back end temperature was used as the process monitor of the various conditions. The special character of this variable is that it can show the normal and abnormal conditions inside the kiln.

In spite of great importance of fuzzy neural networks for solving wide range of real-world problems, unfortunately, little progress has been made in their development.

We have discussed recurrent neural networks with fuzzy weights and biases as adjustable parameters and internal feedback loops, which allows capturing dynamic response of a system without using external feedback through delays. In this case all the nodes are able to process linguistic information.

As the main problem regarding fuzzy and recurrent fuzzy neural networks that limits their application range is the difficulty of proper adjustment of fuzzy weights and biases, we put an emphasize on the TNFS training algorithm.

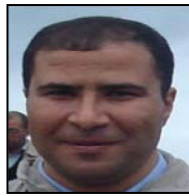

**First Authorname**
MAHDAOUI Rafik holds the engineer Diploma from the Computer Science department, Hadj LAkhdhar University of Batna, Algeria in 2001; he obtained the degree of Magister in 2008 at Batna University in Industrial engineering. Since 2009 he is an assistant Professor at Computer Science department of Khenchela university Algeria , where he teaches: Programming languages, graph theory, script languages and others matters. He supervises engineers and masters students on their final projects. he is a member of secure operating systems Group, at LAP Laboratory, Hadj LAkhdhar University of Batna, where he is preparing a PHD diploma. he's research interests are in Neuro-Fuzzy systems , Artificial intelligence, emergent technologies , prognosis and diagnosis ,e-maintenance.